\documentclass[runningheads]{llncs}
\usepackage[T1]{fontenc}

\usepackage{graphicx}
\usepackage{amsmath}
\usepackage{amssymb}
\usepackage{booktabs}
\usepackage{placeins}   
\usepackage{hyperref}
\begin{document}
\title{ARC-CT: Anatomy-Routed Contrastive Vision-Language Learning for 3D Chest CT}
\titlerunning{Anatomy-Routed Contrastive Learning for 3D Chest CT}
\authorrunning{H. U. Isik et al.}
%
\author{Huseyin Umut Isik\inst{1}\thanks{Equal contribution.\quad
$^{\star\star}$Equal contribution as senior authors.\quad
$^{\dagger}$Corresponding author: \email{alp.ozaydin@metu.edu.tr}} \and
Mehmet Alp Ozaydin\inst{1}$^{\star,\dagger}$  \and
\newline
Sila Kurugol\inst{3}$^{\star\star}$ \and
\c{S}eyda Ertekin\inst{1,2}$^{\star\star}$}

\institute{Department of Computer Engineering, METU, Ankara, Turkey \and
METU-DTX Digital Transformation and Innovation Center, Ankara, Turkey \and
Quantitative Intelligent Imaging Lab, Boston Children's Hospital and Harvard Medical School, Boston, MA, USA}
%
\maketitle
\begin{abstract}
Contrastive vision-language learning uses paired chest CT volumes and
radiology reports to learn abnormality classifiers without manually annotated labels.
However, two characteristics of chest CT challenge conventional global
contrastive learning. First, many critical abnormalities are
small or anatomically localized, and pooling an entire volume into a single
embedding may dilute their visual evidence. Second, the standard contrastive objective treats every other scan in a batch as a negative. Because many chest CTs share abnormalities, this objective incorrectly pushes co-positive pairs apart. We propose Anatomy-Routed Contrastive Learning for 3D Chest CT (ARC-CT), a region-aware framework that addresses these limitations using only labels extracted from reports by an LLM, with no manual annotations or bounding boxes. ARC-CT combines three components: (1) an AnatomyQFormer localizing evidence via queries constrained by automatically generated organ masks; (2) a label-Jaccard soft InfoNCE objective integrating the standard one-hot target with the label-set overlap of each pair, which reduces false-negative penalties between studies that share clinical findings; and (3) an organ-level alignment loss connecting mask-pooled visual features to organ-specific
report text extracted offline with a large language model. ARC-CT achieves a 0.86 mask-free macro AUC across 18 abnormalities using a compact 3D ResNet-18 backbone. Overall, ARC-CT outperforms both comparable efficient baselines and several larger transformer models. Our code and weights are available at \url{https://github.com/arc-ct/arc-ct}.
\end{abstract}
\section{Introduction}
Chest CT is the primary modality for thoracic disease, and most clinical volumes come with a free-text radiology report. Reports are written as
routine care while per-voxel or per-finding labels are not, so reports exist at
hospital-archive scale while curated labels do not. Contrastive
vision-language pretraining~\cite{clip,convirt,gloria} exploits this pairing: it
associates volumes with reports and then classifies abnormalities from text
prompts. CT-CLIP~\cite{ctclip} established this setup on CT-RATE, a public
dataset of chest CT volumes, reports, and 18 abnormality labels. Follow-up work
has pushed accuracy and efficiency further: MPS-CT~\cite{mpsct} derives
silver-standard labels from reports with an LLM and reaches strong accuracy with
a compact 3D ResNet backbone, GreenRFM~\cite{greenrfm} targets resource
efficiency, and BrgSA~\cite{brgsa}, Merlin~\cite{merlin}, and BIUD~\cite{biud} add
structured, multi-task or distilled signals on larger encoders.
HLIP~\cite{hlip} scales language-image pretraining to volumetric CT, and
ViSD-Boost~\cite{visdboost} raises vision semantic density with anatomy-normality
modeling. fVLM~\cite{fvlm}
decomposes volumes into anatomical regions, aligns them with report
sentences, and reduces false negatives by matching anatomically. However, it
relies on region-level decomposition and a large 3D Swin backbone.
MedCLIP~\cite{medclip} replaces the hard 2D contrastive target with a soft
semantic-similarity target so that clinically similar pairs are not treated as
false negatives.

Two remaining problems are specific to how 3D chest CT is pooled and contrasted, and neither is fully solved by prior work. First, clinically important findings are often small and focal. Pooling an entire 3D volume into a single embedding averages localized evidence, such as a nodule or a patch of consolidation, against thousands of silent voxels. Second, a conventional one-hot contrastive loss introduces false negatives. This objective incorrectly pushes co-positive pairs apart as many scans in a typical batch share the same abnormalities.

ARC-CT is a region-aware contrastive framework that targets both problems directly. Our contributions are:
\begin{enumerate}
\item \textbf{An AnatomyQFormer with role-typed queries (anatomy, pathology, and global) to localize evidence.} Anatomy is injected by masking each anatomy query to its organ region, so no bounding-box or region-proposal supervision is needed. ARC-CT uses a single global image-text objective and folds anatomy in through masked-attention routing and an auxiliary per-organ alignment term rather than separate region encoders.
\item \textbf{A label-Jaccard soft target combined with per-organ alignment.} This approach mitigates the false-negative problem among co-positive scans and introduces region-level supervision, all without requiring language-model inference during training or testing at run time.
\end{enumerate}

These two components, combined with a weakly supervised warm-start, create a compact and highly accurate system. Using a 3D ResNet-18~\cite{r3d} backbone trained on 47K volumes, ARC-CT achieves a mask-free macro AUC of 0.855 on CT-RATE (three-seed mean). It outperforms recent baselines at the same data scale, including models that share the same backbone.

\section{Method}
\begin{figure}[t]
\centering
\includegraphics[width=0.92\textwidth]{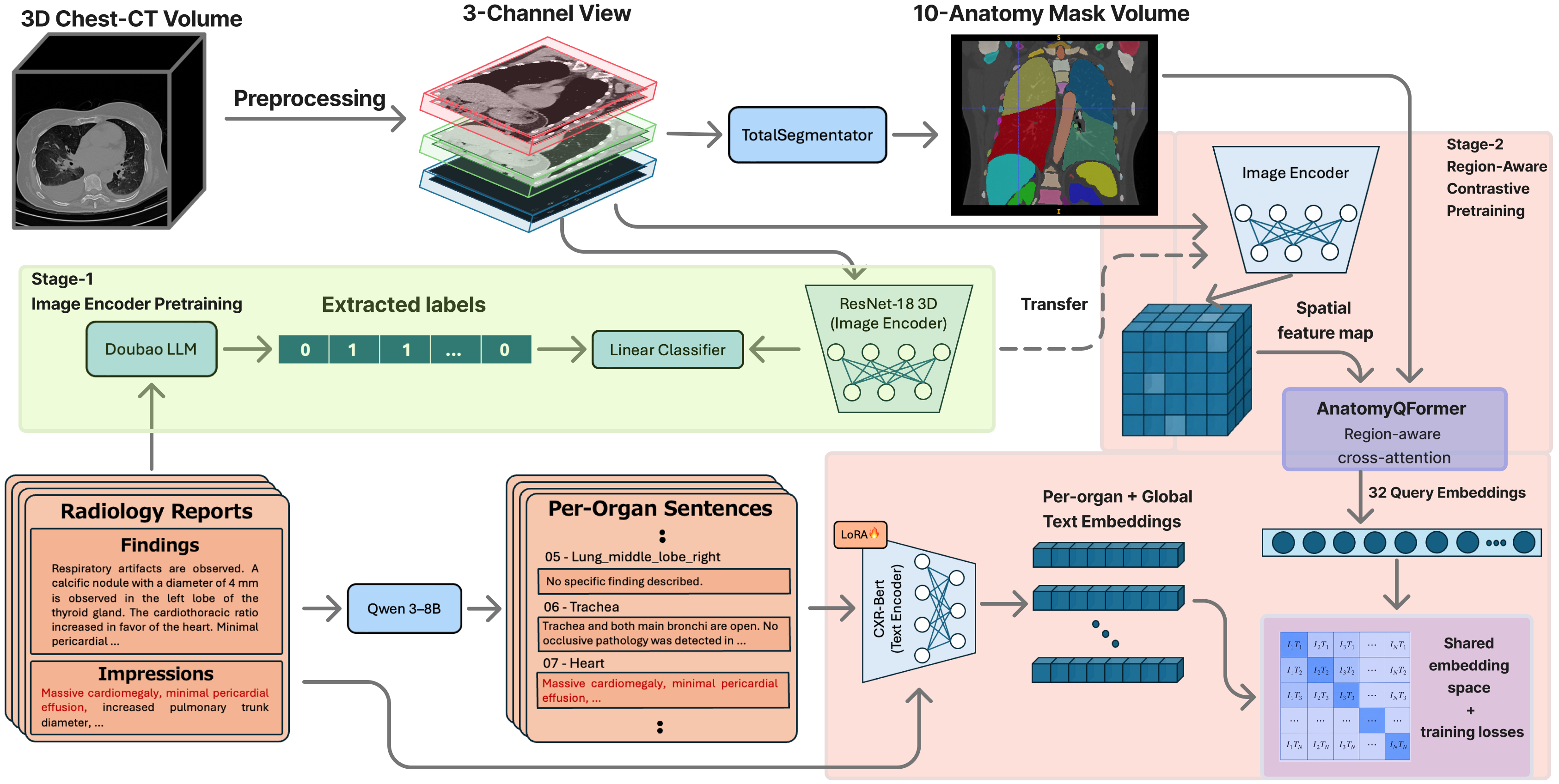}
\caption{\textbf{ARC-CT overview.} Stage 1 pretrains a 3D ResNet-18 on 18 LLM-extracted abnormality labels, warm-starting Stage 2. 
In Stage 2, the encoder's feature map is cross-attended by the AnatomyQFormer's anatomy, pathology, and global queries, while an LLM parses reports into per-organ sentences. Embeddings are projected into a shared space and trained with a label-Jaccard soft InfoNCE contrastive loss with per-organ alignment and query supervision.}
\label{fig:pipeline}
\end{figure}

ARC-CT is trained in two stages. Stage~1 supervises the image encoder on the
18 extracted abnormality labels. Stage~2 performs region-aware contrastive
vision-language pretraining. At inference, ARC-CT classifies each abnormality by comparing the test CT volume against text prompts and requires no masks, as detailed in the Inference paragraph below. Figure~\ref{fig:pipeline} shows the full
pipeline.

\smallskip\noindent\textbf{Input representation.}
Volumes are resampled to $1.5{\times}1.5{\times}3.0$\,mm spacing and cropped to $3{\times}96{\times}192{\times}192$ voxels. Three channels represent clinical Hounsfield-unit (HU) windows exposing complementary tissues: lung $[-1500, 500]$, soft-tissue $[-160, 240]$, and bone $[300, 2000]$.

\smallskip\noindent\textbf{Stage~1: Weakly supervised image pretraining.}
The 3D ResNet-18 image encoder, initialized from Kinetics~\cite{kinetics}, uses global pooling and a linear head for 18 abnormality logits. To handle CT-RATE's class imbalance, we train with asymmetric loss~\cite{asl} ($\gamma_-{=}4$, $\gamma_+{=}1$, $m{=}0.05$) to down-weight easy negatives. Trained on the 47,149-volume training split, this stage reaches a validation AUC of 0.88, serving as a weakly supervised reference point. The linear head is then discarded, and the backbone warm-starts Stage 2.

\smallskip\noindent\textbf{Stage~2: Region-aware contrastive pretraining.}
The image encoder produces a spatial feature map of shape $[512, 12, 12, 12]$, reflecting the Kinetics-style R3D-18 backbone's anisotropic stem stride $(1, 2, 2)$, which downsamples height and width by 16× and depth by 8× overall.
The report is encoded by CXR-BERT~\cite{cxrbert} adapted with LoRA~\cite{lora}
(rank 8 on the query and value projections of the top transformer layers), so
the text tower is almost entirely frozen. This stage combines three components (an
AnatomyQFormer, a label-Jaccard soft InfoNCE objective, and a per-organ alignment
loss) together with an auxiliary query-supervision term, detailed below:

\smallskip\textbf{AnatomyQFormer.} Unlike prior 3D medical Q-Formers~\cite{medblip}, ours routes attention by organ mask without regional cropping. A query transformer~\cite{blip2} cross-attends to the
image feature map with 30 learned queries of three roles: 10 anatomy queries,
18 pathology queries, and 2 global queries. Query $q_o$ is bound to organ $o$ by a fixed index over the five lung lobes,
trachea, heart, aorta, mediastinal vessels, and esophagus. Binding is
attention scope only, with no organ embedding and no per-query label. Each pathology query is restricted to the union of organs its finding can occupy; medical
material, which has no anatomical prior, attends freely. Global queries summarize the volume.

\smallskip\textbf{Label-Jaccard soft InfoNCE.} The core objective is image-text
contrastive learning between the global image and report embeddings $\ell_i$ and
$r_i$. Standard one-hot InfoNCE~\cite{infonce} erroneously penalizes co-positive batch scans. To correct this, we adopt a soft contrastive target~\cite{medclip} that weights each pair by the Jaccard overlap of their label sets, combined with the identity match using a mixing weight $\alpha=0.3$ ($\alpha=0$ recovers one-hot InfoNCE). The matched
pair keeps full weight, while a scan that shares all of scan $i$'s labels is
pulled up rather than pushed away as a negative. We write
$\mathcal{L}_{\mathrm{con}}(\{u_i\},\{v_i\})$ for this soft-InfoNCE loss on a
batch of paired embeddings, so the image-text term is
$\mathcal{L}_{\mathrm{con}}(\{\ell_i\},\{r_i\})$.

\smallskip\textbf{Per-organ alignment.} For each volume, we pool the image feature map inside organ masks to obtain per-organ image embeddings, and pair them with the corresponding report sentences, parsed offline by Qwen3-8B~\cite{qwen}. A second soft InfoNCE loss aligns these image-text pairs, encouraging the anatomy queries to encode organ-specific content. Reusing the soft-InfoNCE form on per-organ embeddings $\ell_i^{o}$ and $r_i^{o}$ for organ $o$ yields

{%

\begin{equation}
\mathcal{L}_{\mathrm{org}}
=
\frac{1}{|\mathcal{O}|}
\sum_{o\in\mathcal{O}}
\mathcal{L}_{\mathrm{con}}
\bigl(\{\ell_i^{o}\},\{r_i^{o}\}\bigr).
\end{equation}
}

\smallskip\textbf{Query supervision.} Each pathology query is supervised against its class prompt with a per-token objective, and for the focal classes (lung nodule, atelectasis, lung opacity, consolidation) the query response is pooled over the most active spatial locations rather than the whole grid, preventing small lesions from being averaged out during training. Each class $c$ carries a positive and a negative text prompt with embeddings $g_c^{+}$ and $g_c^{-}$, and the query is scored by a two-way softmax over the two prompts followed by binary cross-entropy against the label:
{%
\begin{equation}
\hat{p}_{ic}
=
\frac{
e^{\hat{\ell}_i^{\top}g_c^{+}/\tau}
}{
e^{\hat{\ell}_i^{\top}g_c^{+}/\tau}
+
e^{\hat{\ell}_i^{\top}g_c^{-}/\tau}
},
\qquad
\mathcal{L}_{\mathrm{cls}}
=
\frac{1}{BC}
\sum_{i=1}^{B}\sum_{c=1}^{C}
\mathrm{BCE}\bigl(\hat{p}_{ic},y_{ic}\bigr).
\end{equation}
}
\noindent The total Stage-2 objective is a weighted sum of the contrastive,
query-supervision, and alignment terms ($\lambda_{\mathrm{con}}=\lambda_{\mathrm{org}}=1$
and $\lambda_{\mathrm{cls}}=0.5$):
{%

\begin{equation}
\mathcal{L}
=
\lambda_{\mathrm{con}}
\mathcal{L}_{\mathrm{con}}
\bigl(\{\ell_i\},\{r_i\}\bigr)
+
\lambda_{\mathrm{cls}}
\mathcal{L}_{\mathrm{cls}}
+
\lambda_{\mathrm{org}}
\mathcal{L}_{\mathrm{org}}.
\end{equation}
}

\smallskip\noindent\textbf{Inference.}
At test time, the AnatomyQFormer does not use organ masks, and all output queries are averaged into a single embedding. Following CT-CLIP~\cite{ctclip}, ARC-CT evaluates this embedding against positive and negative text prompts (e.g., ``Cardiomegaly'' vs.\ ``No Cardiomegaly''). The score is the softmax weight on the positive prompt, which requires no segmentation masks or LLM calls. Supplying masks at inference changes macro AUC by only $+0.001$, which shows that mask-free inference causes no meaningful performance loss.

\section{Experiments and Results}
\subsection{Dataset and labels}
CT-RATE~\cite{ctclip} is a public non-contrast chest CT dataset. It consists of 50{,}188
reconstructed volumes with paired reports from 21{,}304 patients. We follow the
official split: 47{,}149 training and 3{,}039 validation volumes. We extract the 18 training labels from training reports using the Doubao~\cite{doubao} LLM, following the protocol of MPS-CT~\cite{mpsct}. For evaluation, we use the official ground-truth labels from the validation set. We generate organ masks using TotalSegmentator~\cite{totalseg} and reduce them to ten groups relevant to thoracic disease: the five lung lobes, trachea, heart, aorta, mediastinal vessels, and esophagus. For external evaluation, we use the public subset of the RAD-ChestCT~\cite{radchestct} dataset (3,630 volumes). We map its labels to the 18 CT-RATE abnormalities following CT-CLIP~\cite{ctclip}. This dataset serves strictly as an external benchmark and is never used for training.

\subsection{Implementation details}
Stage~1 trains the image encoder for 8 epochs (batch size 8, learning rate $1\times10^{-4}$) using asymmetric loss. Stage~2 trains for 14{,}400 updates (1{,}200 warm-up) at batch size 20, using a vision learning rate of $1.1\times10^{-5}$ and a smaller text-LoRA rate. We optimize via AdamW with bf16 mixed precision on one 80\,GB A100 GPU. The image encoder and projection layers remain trainable, while the text encoder is frozen except for its LoRA parameters. The learnable temperature is initialized to 0.07, and augmentation is limited to HU jitter. Hyperparameters were tuned on a 2{,}000-volume subset of the
training split. Inference takes $\sim$35\,ms per volume with 2.0\,GB peak memory at batch size 1. Our code and trained weights are publicly available at \url{https://github.com/arc-ct/arc-ct}.

\subsection{Main results}
\begin{figure}[t]
\centering
\includegraphics[width=0.9\textwidth]{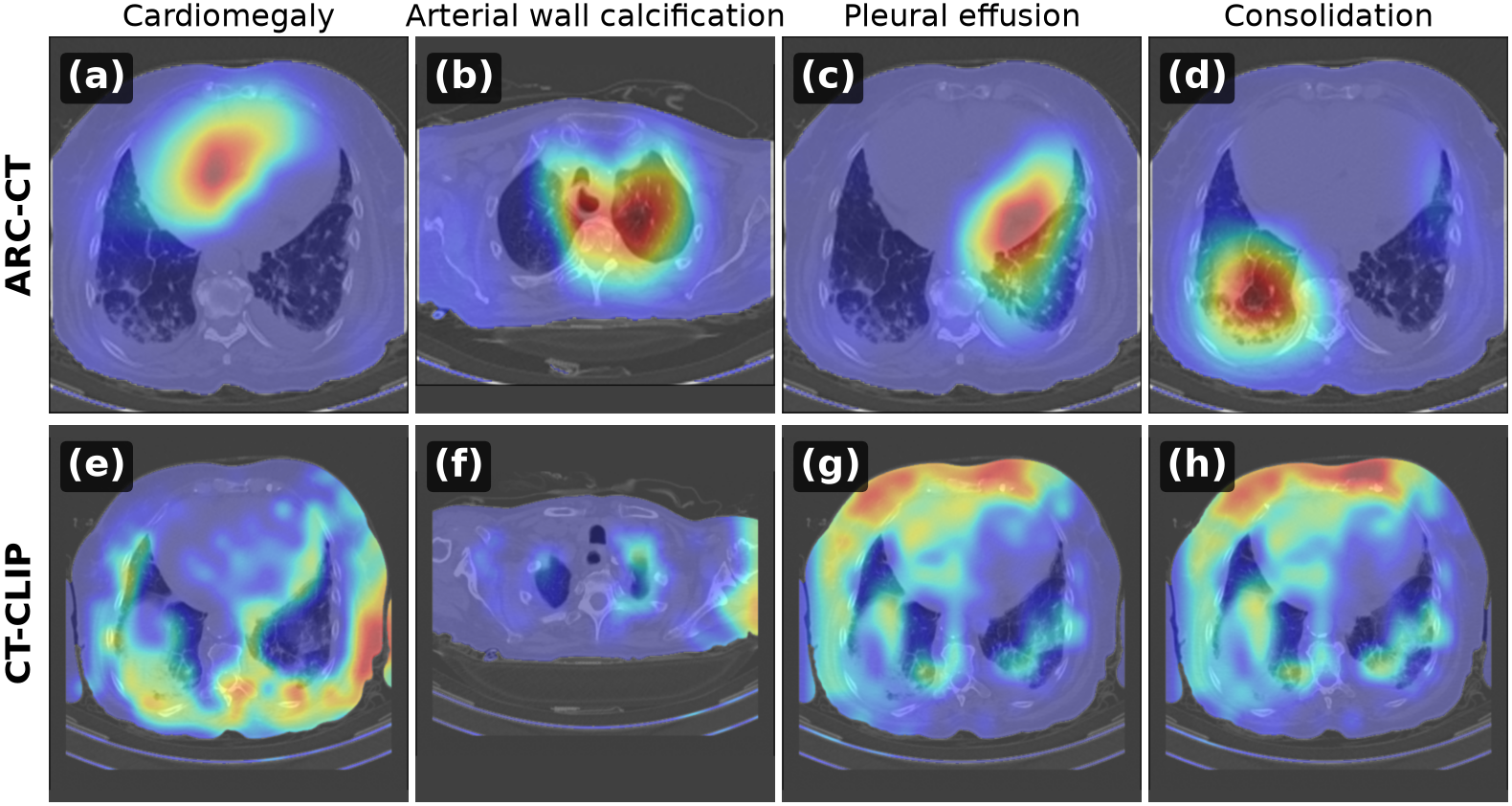}
\caption{Mask-free Grad-CAM for four positive findings (cardiomegaly, arterial wall calcification, pleural effusion, consolidation) on one CT-RATE validation
volume. Top
row (a--d): ARC-CT; bottom row (e--h): CT-CLIP~\cite{ctclip}. ARC-CT's
activation is sharply localized to the relevant structure, while CT-CLIP's is
diffuse and largely unchanged between findings.}
\label{fig:attn}
\end{figure}

Table~\ref{tab:main} compares ARC-CT with published methods. ARC-CT reaches a macro AUC of 0.855 (three-seed mean), surpassing MPS-CT~\cite{mpsct} at the same data scale and
GreenRFM~\cite{greenrfm} despite its larger training set; it also exceeds larger
transformer-based methods such as BrgSA~\cite{brgsa} and fVLM~\cite{fvlm}. ARC-CT leads on accuracy (0.787), F1 (0.809), and precision (0.455). On external RAD-ChestCT, ARC-CT achieves a macro AUC of 0.734 without fine-tuning. Despite requiring no bounding-box supervision, Grad-CAM~\cite{gradcam} maps (Figure~\ref{fig:attn}) show ARC-CT sharply localizes findings, whereas CT-CLIP yields diffuse activation.

\begin{table}[t]
\caption{Prompt-based abnormality classification (no manual labels) on CT-RATE (in-domain) and
RAD-ChestCT (external). The best per column in \textbf{bold}, second
\underline{underlined}.}\label{tab:main}
\centering
\setlength{\tabcolsep}{4pt}
\footnotesize
\begin{tabular}{lcccccccc}
\toprule
& \multicolumn{4}{c}{CT-RATE (in-domain)} & \multicolumn{4}{c}{RAD-ChestCT (external)} \\
\cmidrule(lr){2-5}\cmidrule(lr){6-9}
Method$^{\ast}$ & AUC & Prec & Acc & F1 & AUC & Prec & Acc & F1 \\
\midrule
\textbf{ARC-CT (ours)} & \textbf{0.855} & \textbf{0.455} & \textbf{0.787} & \textbf{0.809} & 0.734 & 0.392 & 0.679 & 0.716 \\
GreenRFM~\cite{greenrfm} & \underline{0.848} & \underline{0.444} & \underline{0.777} & \underline{0.800} & \textbf{0.780} & \textbf{0.461} & \textbf{0.730} & \textbf{0.757} \\
MPS-CT~\cite{mpsct} & 0.838 & 0.435 & 0.774 & 0.796 & \underline{0.773} & \underline{0.450} & \underline{0.719} & \underline{0.748} \\
BrgSA~\cite{brgsa} & 0.792 & 0.385 & 0.733 & 0.762 & 0.742 & 0.422 & 0.686 & 0.720 \\
ViSD-Boost~\cite{visdboost} & 0.790 & 0.387 & 0.731 & 0.759 & 0.694 & 0.342 & 0.652 & 0.693 \\
HLIP~\cite{hlip} & 0.787 & 0.384 & 0.724 & 0.755 & 0.717 & 0.398 & 0.677 & 0.714 \\
fVLM~\cite{fvlm} & 0.778 & 0.379 & 0.718 & 0.751 & 0.680 & 0.374 & 0.647 & 0.688 \\
CT-CLIP~\cite{ctclip} & 0.731 & 0.323 & 0.668 & 0.707 & 0.629 & 0.336 & 0.595 & 0.642 \\
Merlin~\cite{merlin} & 0.728 & 0.337 & 0.672 & 0.709 & 0.644 & 0.348 & 0.619 & 0.663 \\
BIUD~\cite{biud} & 0.713 & 0.338 & 0.681 & 0.716 & 0.629 & 0.337 & 0.606 & 0.652 \\
\bottomrule
\end{tabular}
\par\smallskip
\parbox{\textwidth}{\footnotesize $^{\ast}$Following the CT-CLIP
protocol~\cite{ctclip}, F1 is averaged with support-weighting while precision is
reported for the positive class only. Baseline numbers are quoted as
published: from~\cite{mpsct} for CT-CLIP, BIUD, Merlin, fVLM, BrgSA, and MPS-CT;
from~\cite{greenrfm} for GreenRFM, HLIP and ViSD-Boost.}
\end{table}

\begin{table}[t]
\caption{Per-class AUC on CT-RATE validation (3{,}039 volumes), three-seed mean.
GreenRFM~\cite{greenrfm}, MPS-CT~\cite{mpsct}, and CT-CLIP~\cite{ctclip} are the
baselines with a published 18-class breakdown. Best per row in \textbf{bold},
second \underline{underlined}.}
\label{tab:perclass}
\centering
\setlength{\tabcolsep}{4pt}
\footnotesize
\begin{tabular}{lcccc}
\toprule
Class & ARC-CT & GreenRFM & MPS-CT & CT-CLIP \\
\midrule
Pleural effusion & \textbf{0.97} & \textbf{0.97} & \underline{0.96} & 0.90 \\
Cardiomegaly & \underline{0.93} & \underline{0.93} & \textbf{0.94} & 0.86 \\
Arterial wall calcification & \textbf{0.94} & \textbf{0.94} & \underline{0.93} & 0.85 \\
Coronary artery wall calcification & \underline{0.93} & \textbf{0.94} & \underline{0.93} & 0.85 \\
Consolidation & \textbf{0.91} & \textbf{0.91} & \underline{0.89} & 0.72 \\
Medical material & \textbf{0.94} & \underline{0.91} & 0.89 & 0.72 \\
Pericardial effusion & \textbf{0.91} & \underline{0.89} & \textbf{0.91} & 0.77 \\
Lung opacity & \textbf{0.88} & \underline{0.85} & 0.83 & 0.63 \\
Mosaic attenuation pattern & 0.86 & \underline{0.88} & \textbf{0.89} & 0.77 \\
Peribronchial thickening & \textbf{0.81} & 0.79 & \underline{0.80} & 0.69 \\
Hiatal hernia & 0.82 & \underline{0.83} & \textbf{0.85} & 0.70 \\
Emphysema & \textbf{0.82} & \underline{0.80} & 0.78 & 0.74 \\
Interlobular septal thickening & \textbf{0.88} & \underline{0.86} & 0.83 & 0.77 \\
Bronchiectasis & \textbf{0.82} & \underline{0.79} & 0.78 & 0.65 \\
Atelectasis & \textbf{0.80} & \underline{0.79} & 0.78 & 0.68 \\
Lymphadenopathy & \textbf{0.78} & \underline{0.77} & \underline{0.77} & 0.70 \\
Lung nodule & \textbf{0.72} & \underline{0.70} & 0.66 & 0.57 \\
Pulmonary fibrotic sequela & \textbf{0.72} & \textbf{0.72} & \underline{0.68} & 0.57 \\
\midrule
\textbf{Macro} & \textbf{0.86} & \underline{0.85} & 0.84 & 0.73 \\
\bottomrule
\end{tabular}
\end{table}

\FloatBarrier

\smallskip\noindent\textbf{Per-class AUC.} Table~\ref{tab:perclass} gives the
per-class breakdown against the baselines that publish one. ARC-CT exceeds
CT-CLIP on every class and matches or beats MPS-CT and GreenRFM on most.
Absolute AUC stays lowest for the small and subtle findings, but that is
where ARC-CT gains most over MPS-CT, by 0.06 on lung nodule and 0.05 on lung
opacity. This is consistent with anatomy routing steering capacity toward
the localized evidence that whole-volume pooling dilutes.

\smallskip\noindent\textbf{Ablations.} Table~\ref{tab:ablation} reports a
three-seed leave-one-out ablation with paired-bootstrap CIs. The weakly
supervised warm-start gives the largest gain, then anatomy routing and the
label-Jaccard target. Per-organ alignment and query supervision are small
but their CIs exclude zero.

\begin{table}[t]
\caption{Leave-one-out ablation on CT-RATE (three-seed mean $\pm$ s.d.; 95\% CIs from a paired patient bootstrap; all $p<0.001$). The weakly supervised Stage-1 warm-start
contributes the largest gain, followed by anatomy routing and the label-Jaccard target.}
\label{tab:ablation}
\centering
\small
\setlength{\tabcolsep}{8pt}
\begin{tabular}{lccc}
\toprule
Configuration & Macro AUC & $\Delta$ & 95\% CI \\
\midrule
 ARC-CT & $0.855 \pm 0.001$ & -- & $[0.848, 0.862]^{}$ \\
\quad $-$ Stage-1 warm-start (Kinetics only) & $0.826 \pm 0.001$ & $-0.028$ & $[-0.032, -0.024]$ \\
\quad $-$ anatomy routing (plain Q-Former) & $0.837 \pm 0.004$ & $-0.017$ & $[-0.019, -0.014]$ \\
\quad $-$ label-Jaccard soft target & $0.845 \pm 0.001$ & $-0.009$ & $[-0.011, -0.007]$ \\
\quad $-$ learnable temperature (fixed) & $0.847 \pm 0.002$ & $-0.007$ & $[-0.009, -0.006]$ \\
\quad $-$ per-organ alignment & $0.850 \pm 0.002$ & $-0.004$ & $[-0.006, -0.002]$ \\
\quad $-$ per-token query supervision & $0.851 \pm 0.001$ & $-0.003$ & $[-0.004, -0.002]$ \\
\bottomrule
\end{tabular}
\end{table}

\smallskip\noindent\textbf{Retrieval.} On cross-modal retrieval
(Table~\ref{tab:retrieval}) ARC-CT leads in both directions: achieving the best or tied-best mean average precision
(MAP) at every image-to-image cut-off and ranking first on report-to-image recall (R@K) at
R@5, R@10, and R@50 except at R@100. Retrieval reuses the same embedding as classification with no
retrieval-specific tuning.

\begin{table}[t]
\caption{Cross-modal image-to-image and report-to-image retrieval on CT-RATE validation, no fine-tuning. ARC-CT leads on all metrics but R@100.
The best per column in \textbf{bold}, second
\underline{underlined}.}
\label{tab:retrieval}
\centering
\setlength{\tabcolsep}{5pt}
\footnotesize
\begin{tabular}{lccccccc}
\toprule
& \multicolumn{3}{c}{Image$\to$image (MAP@$K$)} & \multicolumn{4}{c}{Report$\to$image (R@$K$)} \\
\cmidrule(lr){2-4}\cmidrule(lr){5-8}
Method & @5 & @10 & @50 & @5 & @10 & @50 & @100 \\
\midrule
\textbf{ARC-CT (ours)} & \textbf{71.3} & \textbf{61.7} & \textbf{54.3} & \textbf{11.4} & \textbf{18.0} & \textbf{40.4} & 52.1 \\
GreenRFM~\cite{greenrfm} & -- & -- & -- & 9.5 & 16.0 & \underline{39.9} & \textbf{54.3} \\
MPS-CT~\cite{mpsct} & \textbf{71.3} & \underline{61.5} & \underline{53.7} & \underline{10.0} & \underline{16.3} & 39.0 & \underline{52.2} \\
BrgSA~\cite{brgsa} & \underline{69.2} & 58.5 & 50.5 & 5.8 & 10.1 & 28.6 & 42.0 \\
CT-CLIP~\cite{ctclip} & 68.3 & 57.2 & 48.9 & 2.9 & 5.0 & 18.0 & 28.7 \\
Merlin~\cite{merlin} & 62.6 & 51.3 & 43.9 & 1.5 & 2.7 & 7.7 & 12.7 \\
\bottomrule
\end{tabular}
\end{table}

\section{Discussion and Conclusion}
\label{sec:disc}
ARC-CT reaches 0.86 macro AUC on CT-RATE with two changes that need no spatial
annotation and no extra supervision: anatomy routing from automatically generated
organ masks, and a label-Jaccard soft target that removes a concrete source of
false-negative gradient. This places ARC-CT ahead of MPS-CT, which shares our backbone, dataset, and a comparable supervised warm-start, and ahead of GreenRFM, despite GreenRFM's larger training set. This performance advantage stems directly from how the volume and report signals are structured, not from the encoder size or warm-start alone.

\smallskip
This structuring also yields clearer visual grounding. The AnatomyQFormer's masked attention routing helps the model generate focused attention maps, as shown in Figure~\ref{fig:attn}. The pathology queries highlight specific lesions instead of spreading attention across the entire volume. Despite these strong in-domain results, ARC-CT places fourth on the external RAD-ChestCT dataset, behind methods trained on more heterogeneous scanner and protocol data. Since ARC-CT learns from a single dataset and labeling pipeline, its organ masks and text supervision, while effective in-domain, have not been tested against this kind of protocol shift. Future work will add multi-source pretraining with reconstruction-kernel and
dose augmentation, and recalibrate prompt thresholds on a small unlabeled
target sample. Routing itself could also drop masks, replacing them with a
spatially regularized slot prior that discovers organ boundaries rather than
inheriting them.

\begin{credits}
\subsubsection{\ackname} We thank Mert Sonmezer and Veronika Spieker for helpful research discussions and for proofreading and feedback on the manuscript.

\subsubsection{\discintname}
The authors have no competing interests to declare that are relevant to the content of this article.
\end{credits}

\FloatBarrier

\end{document}